\documentclass[10pt,twocolumn,letterpaper]{article}

\usepackage{authblk}

\usepackage[pagenumbers]{cvpr} %

\usepackage{graphicx}
\usepackage{amsmath}
\usepackage{amssymb}
\usepackage{booktabs}
\usepackage{comment}

\usepackage{float}
\usepackage[font={small}]{caption}
\usepackage[font={small}]{subcaption}

\usepackage{multirow}
\usepackage[utf8]{inputenc} %
\usepackage[T1]{fontenc}    %
\usepackage{makecell}

\usepackage[dvipsnames]{xcolor}

\usepackage{pifont}

\usepackage[pagebackref=true,breaklinks=true,letterpaper=true,colorlinks,bookmarks=false]{hyperref}

\usepackage[capitalize]{cleveref}
\crefname{section}{Sec.}{Secs.}
\Crefname{section}{Section}{Sections}
\Crefname{table}{Table}{Tables}
\crefname{table}{Tab.}{Tabs.}

\def\cvprPaperID{5} %
\def\confName{CVPR}
\def\confYear{2022}

\newcommand{\feat}{\mathbf{F}}%
\newcommand{\img}{\mathbf{I}}%
\newcommand{\graph}{\mathcal{G}}%
\newcommand{\vertice}{\mathcal{V}}%
\newcommand{\edge}{\mathcal{E}}%

\newcommand{\aff}{\mathcal{A}}%

\newcommand{\symmask}{\mathbf{M}_{joint}}%

\newcommand{\predmask}{\mathbf{M}}%

\newcommand{\predcorr}{\mathbf{C}}%

\newcommand{\gtmask}{\mathbf{M}_{gt}}%

\newcommand{\gtcorr}{\mathbf{C}_{gt}}%

\newcommand{\negpool}{N_{pool}}%
\newcommand{\negiter}{K_{hard}}%
\newcommand{\nbeig}{N_{eig}}%
\newcommand{\nbcluster}{K_{cluster}}%

\newcommand{\ce}{CE} %

\newcommand{\lm}{\mathcal{L}_{m}}%
\newcommand{\ltm}{\mathcal{L}_{tm}}%
\newcommand{\lcorr}{\mathcal{L}_{corr}}%
\newcommand{\lsup}{\mathcal{L}}%

\newcommand{\score}{\mathcal{S}}%

\newcommand{\cmark}{\textcolor{ForestGreen}{\text{\ding{51}}}}
\newcommand{\xmark}{\textcolor{red}{\text{\ding{55}}}}

\newcommand{\xishen}[1]{\textcolor{blue}{#1}\PackageWarning{TODO:}{#1!}}

\begin{document}
	
\title{Learning Co-segmentation by Segment Swapping for Retrieval and Discovery}

\author[1]{Xi Shen}
\author[2]{Alexei A. Efros}
\author[3]{Armand Joulin}
\author[4]{Mathieu Aubry}

\affil[1,4]{LIGM (UMR 8049), \'Ecole des Ponts ParisTech}
\affil[2]{University of California, Berkeley}
\affil[3]{Facebook AI Research}

\makeatletter
\renewcommand\AB@affilsepx{, \protect\Affilfont}
\makeatother

\maketitle
	
	\begin{abstract}
		The goal of this work is to efficiently identify visually similar patterns in images, e.g. identifying an artwork detail copied between an engraving and an oil painting, or recognizing parts of a night-time photograph visible in its daytime counterpart. Lack of training data is a key challenge for this co-segmentation task. We present a simple yet surprisingly effective approach to overcome this difficulty: we generate synthetic training pairs by selecting 
		segments in an image 
		and copy-pasting them into another image. We then learn to predict the repeated region 
		masks. We find that it is crucial to predict the correspondences as an auxiliary task and to use Poisson blending and style transfer on the training pairs to generalize on real data. 
		We analyse results with two deep architectures relevant to our joint image analysis task: a transformer-based architecture and Sparse Nc-Net, a recent network designed to predict coarse correspondences using 4D convolutions.
		We show our approach provides clear improvements for artwork details retrieval on the Brueghel dataset and achieves competitive performance on two place recognition benchmarks, Tokyo247 and Pitts30K. We also demonstrate the potential of our approach for unsupervised image collection analysis by introducing a spectral graph clustering approach to object discovery and demonstrating it on the object discovery dataset of~\cite{rubinstein2013unsupervised} and the Brueghel dataset. Our code and data are available at \url{http://imagine.enpc.fr/~shenx/SegSwap/}.
	\end{abstract}
	
	\section{Introduction}
	
	Identifying repeated patterns lies at the very heart of the computer vision problem, and is a key component of Intelligence itself. 
	Yet, in practice, our best methods for performing such a fundamental task often leave a lot to be desired.  
	While we now have good methods for discovering {\em exact} pattern matches (used extensively to find copyright infringements), as well as approximate matches of {\em salient} objects (see object discovery and co-segmentation approaches in Section~\ref{sec:2}), detecting visually similar details within a larger visual context remains surprisingly difficult.   
	
	\begin{figure}[!t]
		\centering
		\subfloat[We blend \xishen{(unsupervised)} image segments and apply style transfer to generate synthetic training data with co-segmentation and correspondence annotations. ]{\includegraphics[width=\columnwidth]{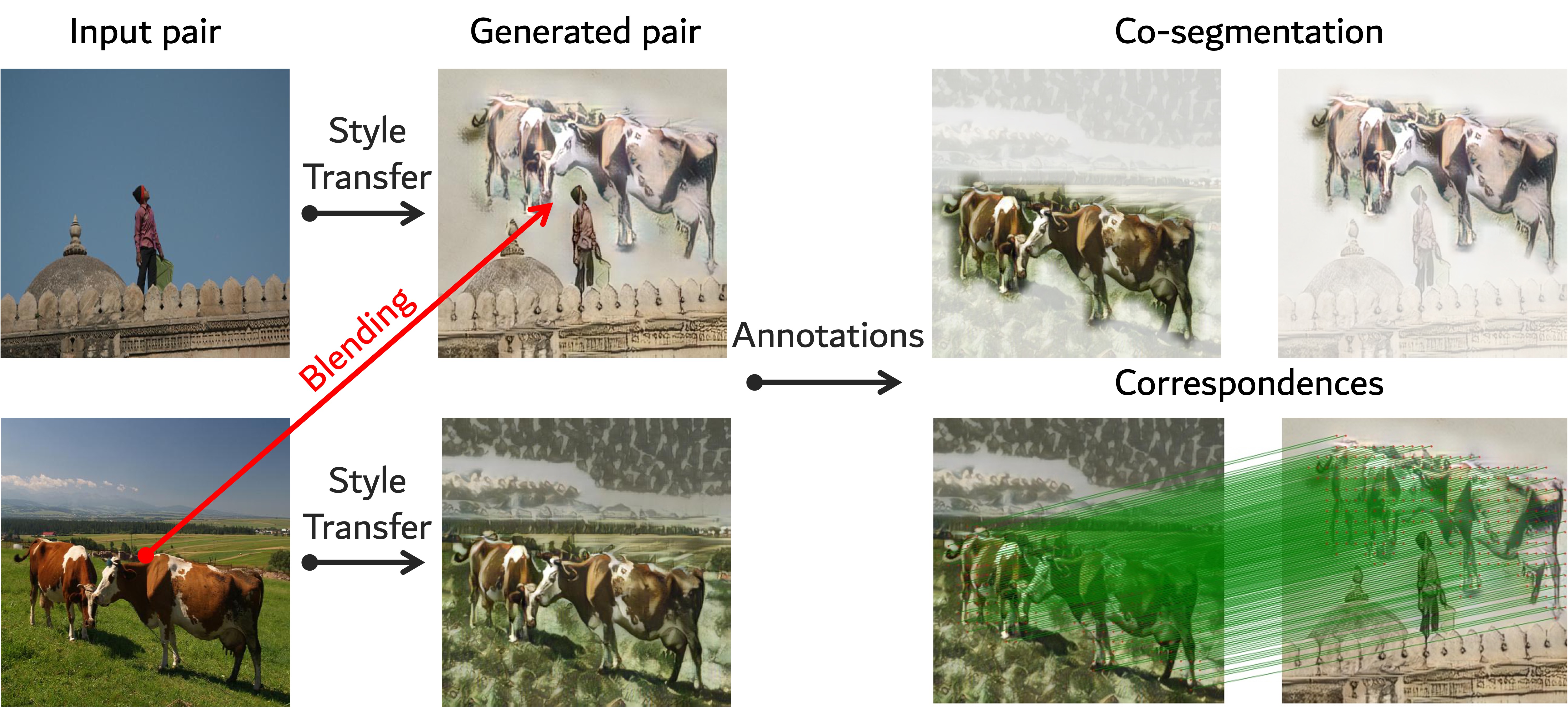}\label{fig:idea_data}}\hfill
		\subfloat[Application on artworks and day/night photograhs]{\includegraphics[width=\columnwidth]{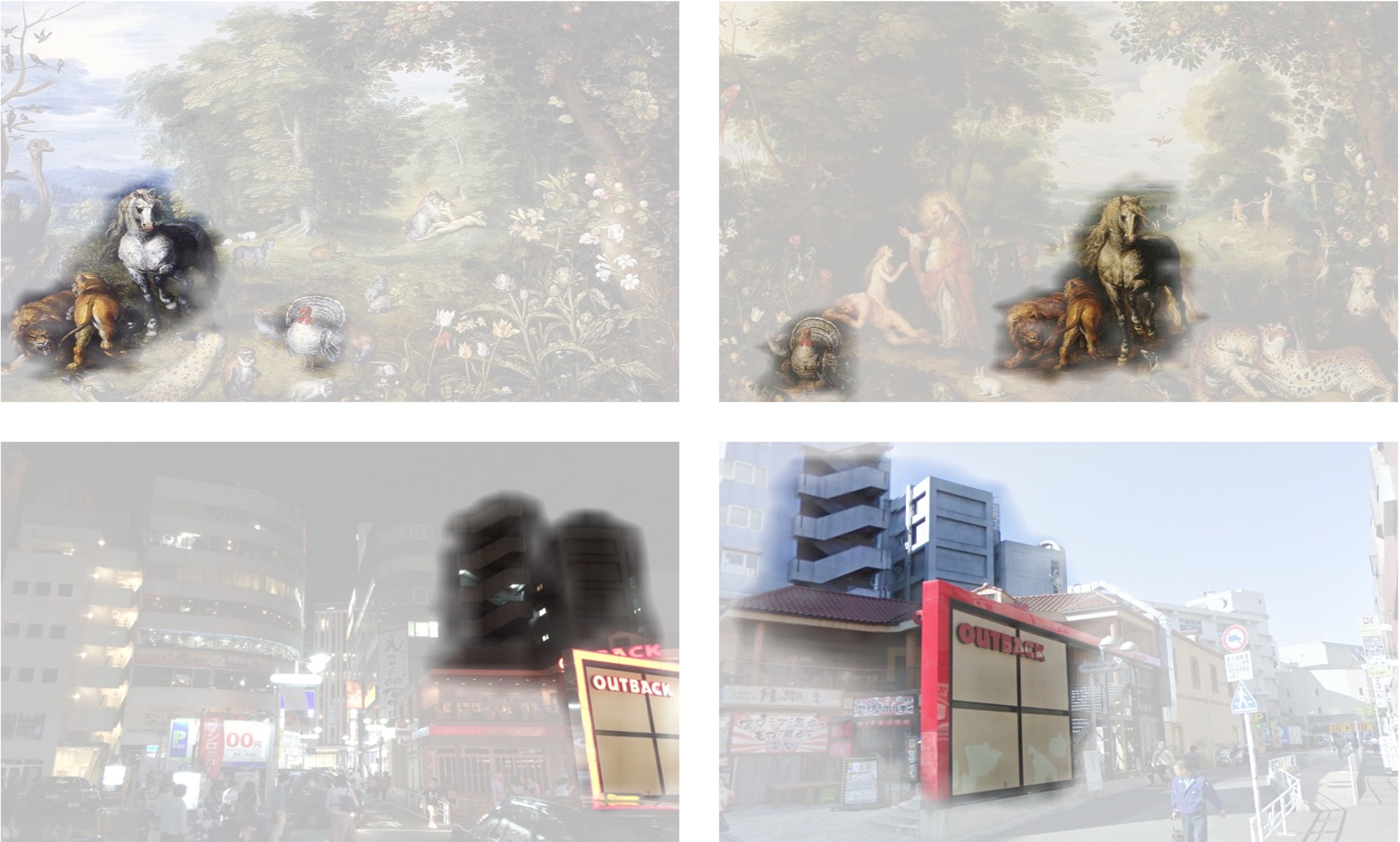}\label{fig:idea_app}}
		\vspace{-3mm}
		\caption{We learn co-segmentation from synthetic images images.
		Models trained on our data are effective for many applications including art detail retrieval, place recognition and object discovery. }
		\label{fig:idea}
	\end{figure}

	Spotting the repetition of visual detail has several applications.  Identifying copied details in artworks allows art historians to discover influences, find provenance, and establish authorship~\cite{shen2019discovering}. Finding repeated details can boost performances in visual localisation for place recognition~\cite{hausler2021patch}. Reliable pair-wise image co-segmentation and correspondence identification could also enable object discovery in image collections~\cite{chen2020show}.

	In this paper, we show it is possible to learn to detect repeated visual patterns -- jointly predicting co-segmentation and correspondences -- without any human-label. 
	{Instead, we generate synthetic  correspondence pairs via automatic data augmentation. More precisely, we use a ``segment swapping'' approach, where we blend image segments in a random background using Poisson blending and apply style transfer to the resulting image to obtain challenging training image pairs (Figure~\ref{fig:idea_data}). On the generated image pairs, we have access to the ground-truth matchability masks as well as the correspondences which we use as supervisions for training a network (Figure~\ref{fig:idea_data}).} Surprisingly, we find that models trained on such a dataset generalize well to real data {and can be directly used for art detail retrieval and place recognition (Figure~\ref{fig:idea_app}).} {We consider using both annotated object segments and unsupervised segments for this data generation process. We obtain unsupervised segments by applying the Bilateral Solver~\cite{barron2016fast} to the recent unsupervised object detection LOST~\cite{LOST}, based on the self-supervised DINO~\cite{caron2021emerging} features, which we show significantly improves performance on unsupervised saliency detection benchmarks.} Experimental results show that using unsupervised segments produces slightly lower but comparable results compared to using COCO~\cite{lin2014microsoft} instance segments.

	We experimented with two network architectures which we adapt to predict co-segmentation and correspondences in image pairs: the recent Sparse Nc-Net ~\cite{rocco2020efficient} architecture, designed for predicting image coarse correspondences, and an architecture based on Transformers~\cite{vaswani2017attention} which we refer to as cross-image transformer. We analyze the effectiveness of our data generation process, architectures and training strategy on two types of tasks. First, we perform retrieval tasks using the predicted pair-wise co-segementation masks and correspondences. We show clear performance improvement for artwork details retrieval on the Brueghel~\cite{shen2019discovering} dataset and results comparable to state of the art for visual localization on two challenging place recognition benchmarks, Tokyo247~\cite{torii201524}  and Pitts30K~\cite{torii2013visual}. This last result is especially impressive, since these benchmarks are very competitive, and many dedicated methods leveraging geo-referenced images or real correspondence for supervision have been proposed. On the contrary, our approach is generic and relies solely on our synthetic ``segment swapping''  training. Second, we make use of the predicted masks and correspondences to build a candidate correspondence graph and introduce an approach to perform discovery in image collections with  spectral clustering~\cite{ng2001spectral,leordeanu2005spectral}. We demonstrate results on par with state-of-the-art on the standard co-segmentation dataset of~\cite{rubinstein2013unsupervised} and show qualitative results on the challenging Brueghel~\cite{shen2019discovering} dataset.  Our code and data are available at \url{http://imagine.enpc.fr/~shenx/SegSwap/}.
	
		\begin{figure}[!t]
		\centering
		\subfloat[Source]{\includegraphics[width=0.48\columnwidth]{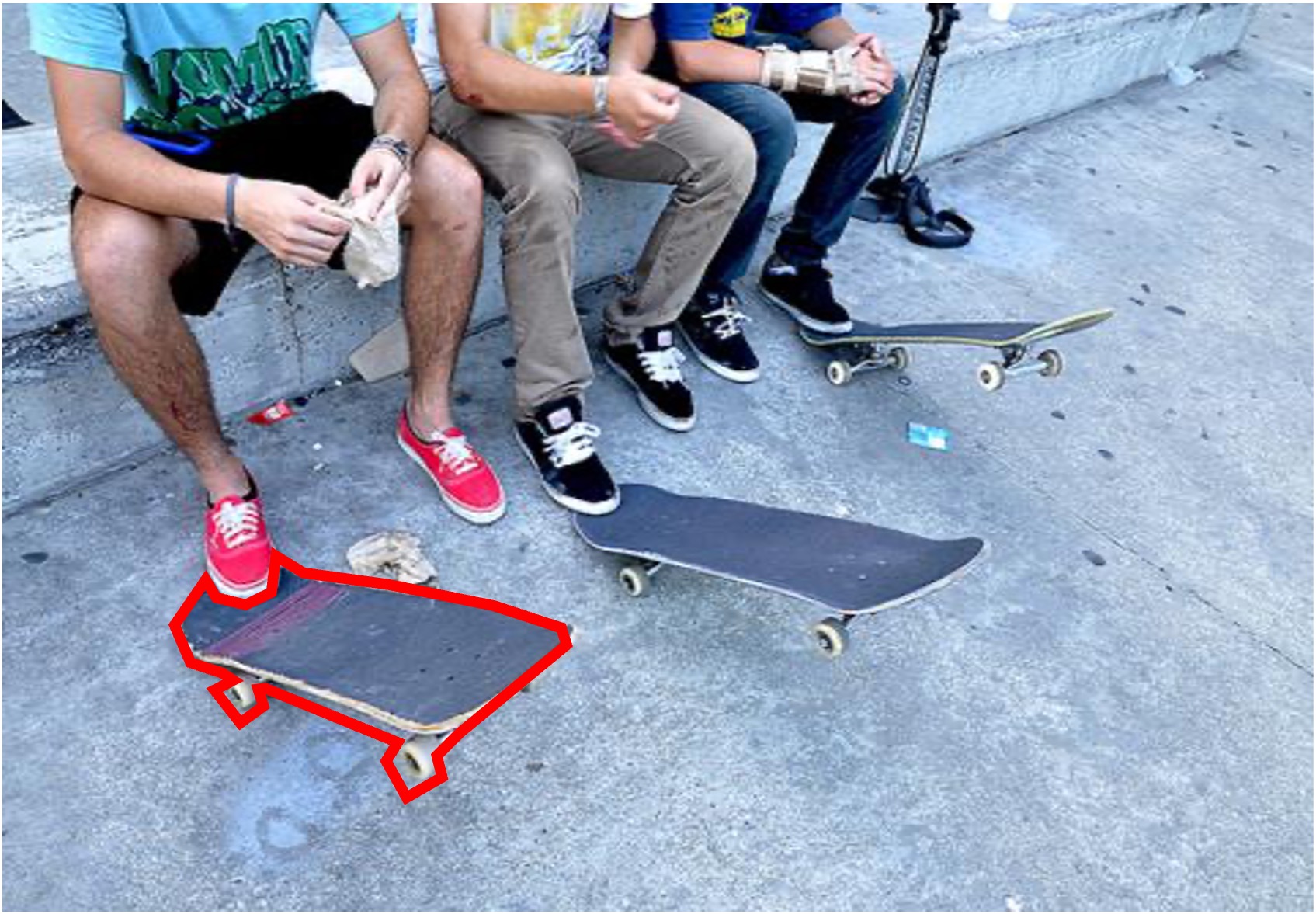}\label{fig:source}}\vspace{2mm}
		\subfloat[Background]{\includegraphics[width=0.48\columnwidth]{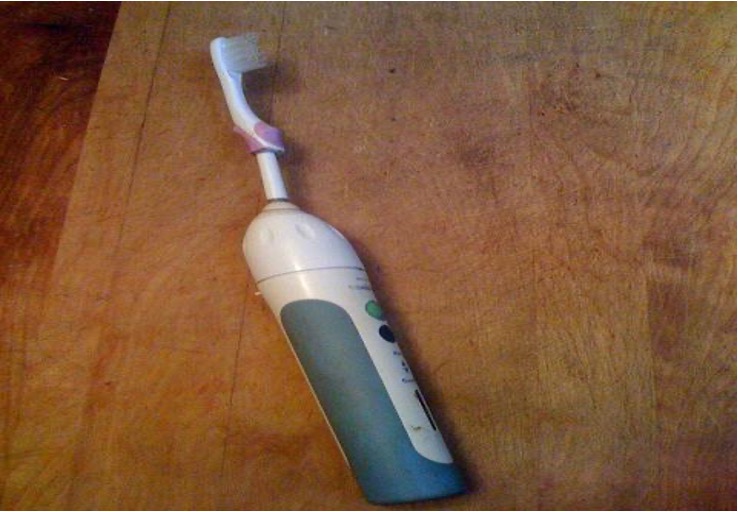}\label{fig:bg}}\hfill
		\subfloat[Direct copy-paste]{\includegraphics[width=0.48\columnwidth]{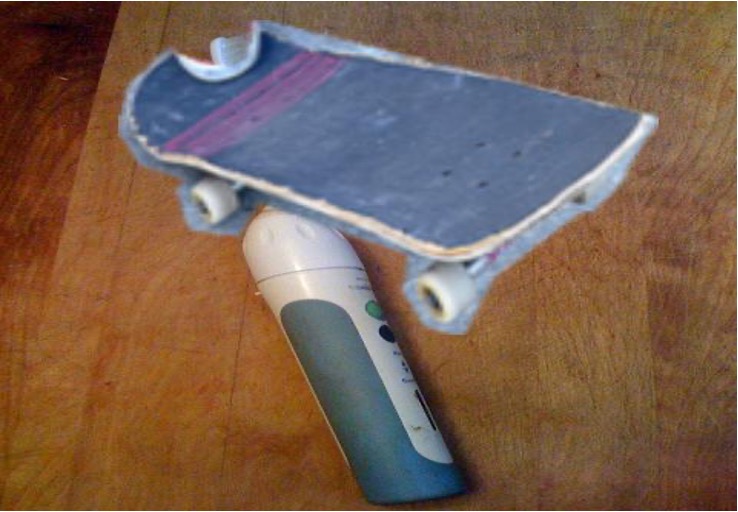}\label{fig:cp}}\vspace{2mm}
		\subfloat[Our blended image]{\includegraphics[width=0.48\columnwidth]{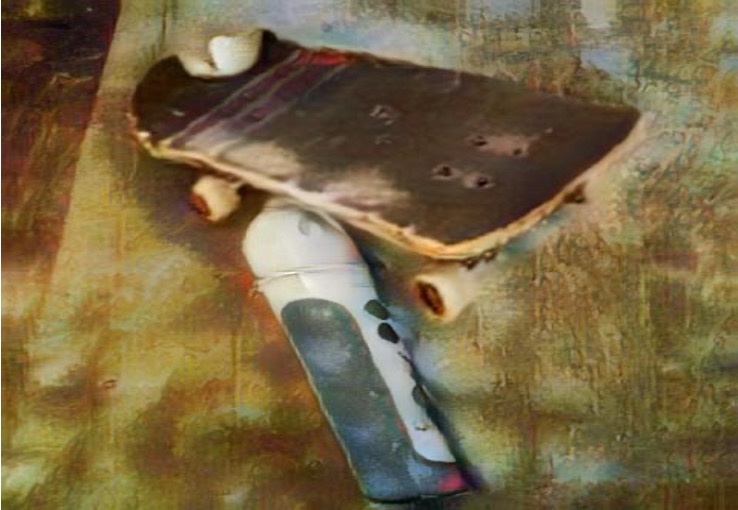}\label{fig:blend}}
		\vspace{-5mm}
		\caption{Data generation by ``segment swapping''. Instead of directly copy-pasting (Figure~\ref{fig:cp}) a segment from a source image (Figure~\ref{fig:source}) to a background (Figure~\ref{fig:bg}), we use Poisson blending~\cite{perez2003poisson} and add style transfer~\cite{huang2017arbitrary} to the result (Figure~\ref{fig:blend}).}
		\label{fig:data}
	\end{figure}
	\begin{figure*}[!thbp]
		\centering
		\includegraphics[width=\textwidth]{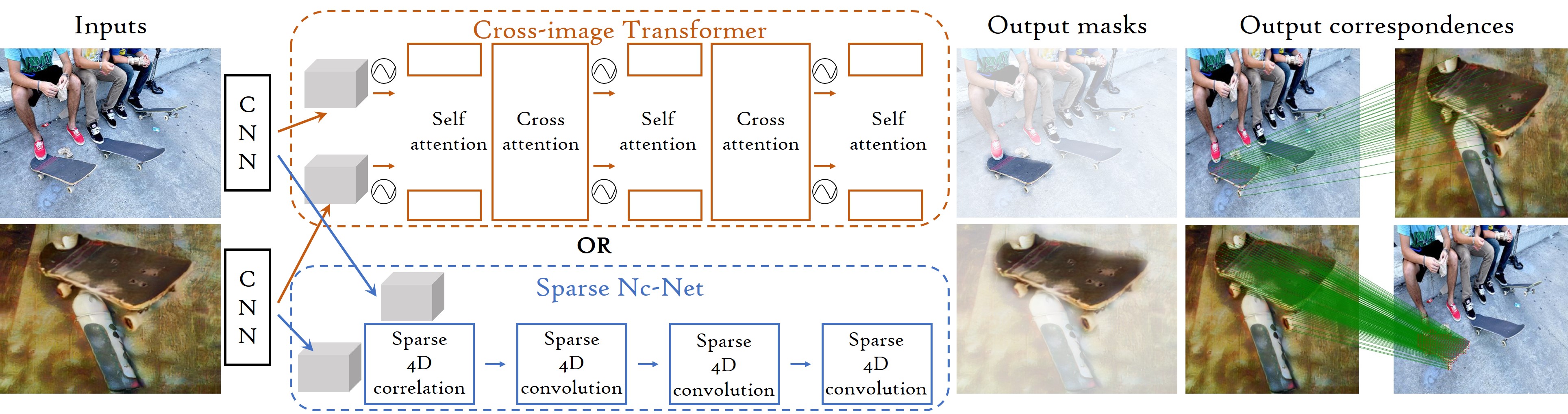}
		\vspace{-7mm}
		\caption{We train our cross-image transformer or Sparse Nc-Net~\cite{rocco2020efficient} on the generated pairs. Both networks jointly predict masks and correspondences.}
		\label{fig:nw}
	\end{figure*}
	\section{Related work}
	\label{sec:2}
	
	\paragraph{Learning correspondences between different images.} 
	SIFT-Flow~\cite{liu2010sift} was an early
	method that aligns visually distinct scenes by incorporating visual features, such as SIFT, into optical flow-style approaches. 
	More recently, many deep learning based approaches have been developed to predict correspondences from correlations of input features~\cite{rocco2017convolutional,melekhov2019dgc,truong2020glu,shen2020ransacflow,truong2021warp,truong2021learning}. 
	Of particular interest, architectures based on attention mechanisms and Transformers~\cite{vaswani2017attention} have been introduced to predict image correspondences. 
	SuperGlue~\cite{sarlin2020superglue} is an attention-based graph neural network for key-point matching. 
	Closer to this work, COTR~\cite{jiang2021cotr} is an sequence-to-sequence transformer architecture that takes an image and 2D coordinates of a query points as inputs to predict correspondences. 
	Finally, LoFTR~\cite{sun2021loftr} adopts a coarse-to-fine approach to matching with a transformer encoder. 
	As opposed to our work, these transformer-based methods are trained on a large dataset with ground-truth poses and depth while we only train on a synthetic dataset. 
	Additionally, our model is only composed of an encoder. More importantly, it outputs an accurate mask of the common regions along with the correspondences. 
	
	\paragraph{Learning correspondences without annotated data.}
	There is a large body of work that use synthetic images~\cite{dosovitskiy2015flownet} or images with synthetic deformations~\cite{rocco2017convolutional,seo2018attentive,melekhov2019dgc,truong2020glu} to learn correspondences without real annotated training data. However, these approaches do not try to identify the matchable regions, which is essential to discover visual details. Some other approaches train directly on real images using proxy signals for 
	correspondences, such as photometric or cycle consistency~\cite{zhou2015flowweb,wang2018occlusion,janai2018unsupervised,shen2020ransacflow,truong2021warp,truong2021learning}.
	Again, they focus on the quality of the correspondences and are not designed to predict matchable regions in vastly different images. 
	On the contrary, the core of our approach is to discover these similar regions.
	This makes our approach particularly suited for retrieval tasks. We are also inspired by related data augmentation techniques, specifically, the CopyPaste augmentation used by Ghiasi~\etal~\cite{ghiasi2020simple} for instance segmentation~ and
	the stylised-ImageNet augmentation used in Geirhos~\etal~\cite{geirhos2018imagenet} to increase shape bias in neural networks.

	\paragraph{Object discovery and co-segmentation.} 	
	
	There is a wide variety of approaches aiming at discovering  objects and their location from unlabelled images. 
	Many methods~\cite{tang2014co,cho2015unsupervised,vo2019unsupervised,vo2020toward} use bounding box proposals and formulate the object discovery as an optimization problem. This relies on the quality of proposals which are typically not adapted for non-photorealistic data, such as artworks. Other approaches~\cite{rother2006cosegmentation,vicente2011object,rubinstein2013unsupervised,taniai2016joint,yuan2017deep,li2018deep,hsu2018co,li2019group,chen2020show} focus on predicting masks of salient objects directly. 
	Some~\cite{yuan2017deep,li2018deep,li2019group} require foreground masks for training, while others~\cite{joulin2010discriminative,joulin2012multi,vicente2011object,hsu2018co,li2019group,chen2020show} are designed to segment common repeated objects in a image collection. 
	These approaches make strong assumptions about the frequency of appearance of an object, while, in many practical scenarios, repeated objects are rare and discovering them is about seeking a needle in a haystack~\cite{shen2019discovering}. Our approach is related to~\cite{rubinstein2013unsupervised,taniai2016joint}, as we both leverage dense correspondences to discover objects. As opposed to our work, Taniai~\etal~\cite{taniai2016joint} focuses on a single pair of images while we also show results over an entire collection of images.
	Rubinstein~\etal~\cite{rubinstein2013unsupervised} makes the assumption that the common object is also the most salient in the image. 
	This works well with images form internet queries but does not apply to artworks where the common object can be a detail in a richer scene.

	\section{Co-segmentation by segment swapping}
	In Section~\ref{sec:data}, we introduce our ``segment swapping'' data generation process (Figure~\ref{fig:data}).  
	We then present in Section~\ref{sec:arch} the two architectures we use (Figure~\ref{fig:nw}). We discuss our loss and training strategy in Section~\ref{sec:loss}. 
	
	\subsection{Training data generation by segment swapping}
	\label{sec:data}

	\paragraph{Training pairs generation} We generate training pairs using images from the COCO dataset~\cite{lin2014microsoft}. We first sample a source image, from which we extract one or two {segments} (as explained below). We then build the target image by applying geometric transformations to the segments and blending them into a random background image using Poisson blending~\cite{perez2003poisson}. The geometric transformations include rotation, translation, scaling, and thin-plate spline (TPS). 
	A style augmentation is then performed on both the source and target images using an AdaIN~\cite{huang2017arbitrary} model trained on the Brueghel dataset~\cite{shen2019discovering}. An example of training pair can be seen in Figure~\ref{fig:data} and we provide more examples of training samples in the supplementary material~\cite{supp_mat}.

	\paragraph{Segments definition}  {The simplest way to define segments for our data generation process is to use annotated object segments. For our experiments, we used the instance annotations from COCO~\cite{lin2014microsoft}. We compare this approach to a completely unsupervised segment extraction, which we defined using the following strategy:
		(1) given an image, we employ LOST~\cite{LOST} to obtain an object segmentation, which is irregular and coarse on the boundary; 
		(2) we refine the object segmentation using the Bilateral Solver~\cite{barron2016fast}.}
	
{We quantitatively validate our unsupervised segments on unsupervised saliency detection benchmarks in Section~\ref{sec:graph_exp}. More details about the unsupervised procedure and example of unsupervised segments and the associated image pairs are available in the supplementary material~\cite{supp_mat}. 
	}

	\subsection{Architectures}
	\label{sec:arch}
	Our networks take as input a source image $\img^s$ and a target image $\img^t$, from which features maps $\feat^s$ and $\feat^t$ of spatial dimension $W \times H$ are extracted by a feature extraction backbone network. These feature maps are then processed either by our cross-image transformer or our modified sparse Nc-Net~\cite{rocco2020efficient} architecture to predict both the masks of the repeated objects in the source and target images, $\predmask^s\in [0,1]^{W\times H}$ and $\predmask^t\in [0,1]^{W\times H}$ respectively, and the correspondences both from source to target $\predcorr^{s \rightarrow t}$ and target to source $\predcorr^{t \rightarrow s}$. $\predcorr^{s \rightarrow t}$ and $\predcorr^{t \rightarrow s}$ are represented as matrices of size $W\times H \times 2$. To simplify notation, we sometime use the masks as continuous 2D functions, which in practice is done by performing bilinear interpolation.

	\paragraph{Cross-image transformer} 
	We built an architecture based on the classic transformer encoder~\cite{vaswani2017attention} which alternates multi-headed attention and fully connected feed-forward networks (FFN) blocks. The FFN blocks contain two layers with a ReLu non-linearity. Similar to~\cite{sarlin2020superglue}, we use two types of attention layers: one is the standard self-attention (SA) layer, the other one is a cross attention (CA) layer where the attention is {only} computed between features from different images. We include the same 2D positional encoding as DeTR~\cite{carion2020end} on top of the feature map before SA. 
	Our transformer alternates these two types of attention layers as shown in Figure~\ref{fig:nw}, with a total of five attention and FFN blocks. Each attention layer has 2 heads and the dimension of the features is 256.  Our last layer is followed by a sigmoid and has three outputs, that we interpret as masks and correspondences for each image. We provide an ablation study of this architecture in the supplementary material~\cite{supp_mat}.
	
	\paragraph{Sparse Nc-Net} Nc-Net~\cite{rocco2020efficient} is designed to learn coarse correspondences under weak supervision. It takes as input the correlations between $\feat^s$ and $\feat^t$, seen as a 4D volume of affinities $\aff_{input} \in \mathbb{R} ^ {W \times H \times W \times H} $, and processes them with 4D convolutions. {The final 4D convolution predicts affinities $\aff_{pred} \in \mathbb{R} ^ {W \times H \times W \times H} $, on which softmax functions are applied in dimensions corresponding to source and target giving $\aff^s_{pred}(i,j,k,l) = \frac{\exp(\aff_{pred}(i,j,k,l))}{\sum_{k,l} \exp(\aff_{pred}(i,j,k,l))}$ and $\aff^t_{pred}(i,j,k,l) = \frac{\exp(\aff_{pred}(i,j,k,l))}{\sum_{i,j} \exp(\aff_{pred}(i,j,k,l))}$. We use the maxima of these affinities as source and target masks, i.e., $\predmask^s(i,j) = \max_{k, l} \aff^s_{pred}(i,j, k, l)$ and $\predmask^t(k,l) = \max_{i,j} \aff^t_{pred}(i,j, k, l)$. Correspondences are obtained with soft-argmax:}

	\begin{equation}
	\scriptsize
		\begin{split}
			\predcorr^s(i,j) &= \left (\sum_{k, l} \frac{k}{W} \aff^s_{pred} (i, j, k, l), \sum_{k, l} \frac{l}{H} \aff^s_{pred} (i, j, k, l)\right )\\
			\predcorr^t(k,l) &= \left (\sum_{i, j} \frac{i}{W} \aff^t_{pred} (i, j, k, l), \sum_{i, j} \frac{j}{H} \aff^t_{pred} (i, j, k, l) \right)
		\end{split}
		\label{eqn:ncnet_corr}
	\end{equation}
	
	Since 4D convolutions are computational heavy, we instead use sparse 4D convolutions with the same architecture as Sparse Nc-Net~\cite{rocco2020efficient}.

	\begin{figure*}[!thbp]
		\centering
		\subfloat[Retrieval results on Brueghel~\cite{shen2019discovering}. {\color{green}green} bounding-boxes are one-shot detection results.]{\includegraphics[width=0.9\textwidth]{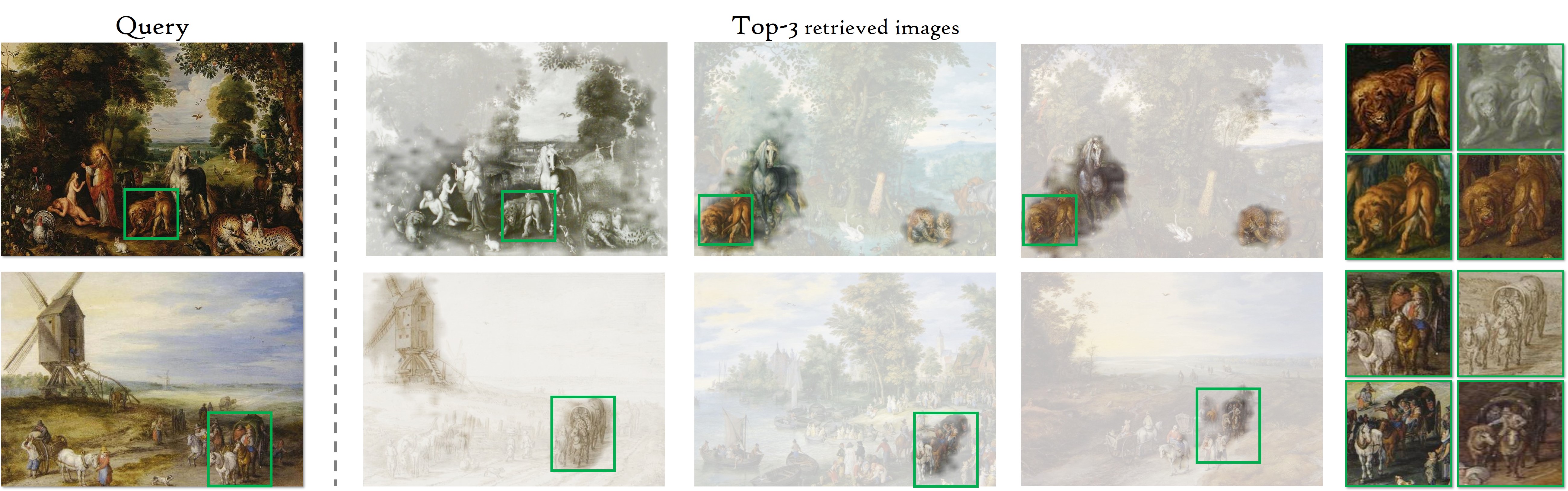}\label{fig:brueghel}} \hfill 
		\subfloat[Retrieval results on Tokyo24/7~\cite{torii201524}.]{\includegraphics[width=0.9\textwidth]{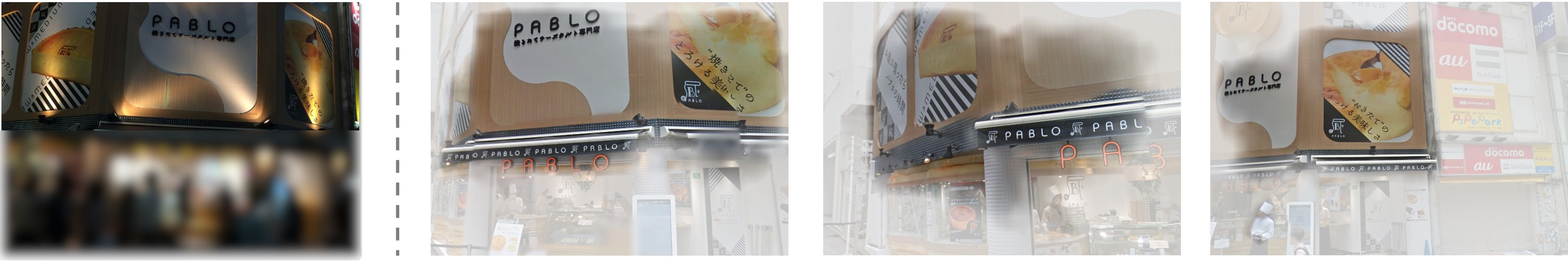}\label{fig:tokyo}} \hfill 
		\subfloat[Retrieval results on Pitts30k~\cite{torii2013visual}.]{\includegraphics[width=0.9\textwidth]{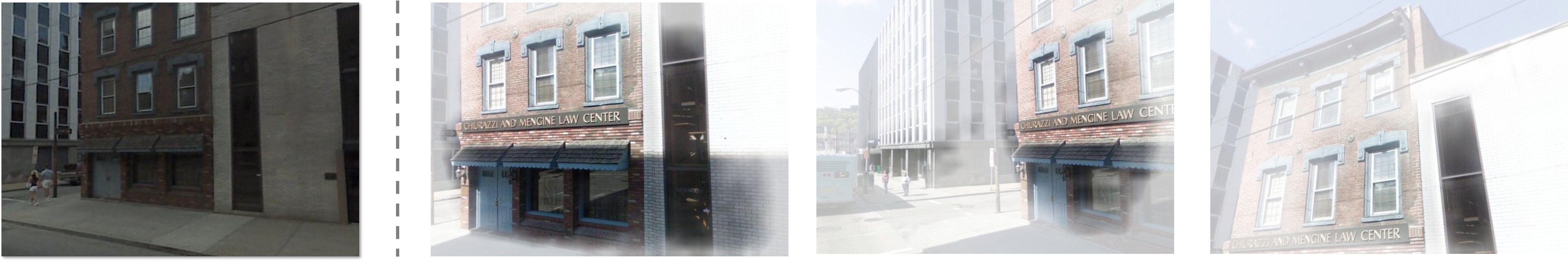}\label{fig:Pitts}}
		\vspace{-1mm}
		\caption{Visual results for retrieval on different datasets. For each query image (1st column), we show its 3 most similar images with the predicted masks as transparency. For Brueghel~\cite{shen2019discovering}, we also show the detection results.}
		\label{fig:vis}
	\end{figure*}
	
	\subsection{Loss and training}
	\label{sec:loss}
	On our synthetic training data we have access to the ground truth masks $\gtmask^s$ and $\gtmask^t$ and ground truth correspondences $\gtcorr^{s \rightarrow t}$ and $\gtcorr^{t \rightarrow s}$ on the source and target images. 
	Our loss is the sum of two symmetric terms for source and target, for simplicity we write only the source loss $\lsup^s$. It includes a cross-entropy ($\ce$) loss on the predicted mask $\lm$ and the transported mask $\ltm$, as well as a regression loss $\lcorr$ on the correspondences: 
	
	\begin{equation}
	    \scriptsize	
		\begin{split}
			\lsup^s  &= \underbrace{\ce(\gtmask^s, \predmask^s)}_{\lm}  + \underbrace{\ce(\gtmask^s,  \predmask^t(\predcorr^{s \rightarrow t} ) )}_{\ltm} \\
			& + \underbrace{\eta \frac{1}{\sum_{i, j} \gtmask^s (i, j)} \sum_{i, j}  \gtmask^s  (i, j) \| \predcorr^{s \rightarrow t}  (i, j) - \gtcorr^{s \rightarrow t}  (i, j) \|}_{\lcorr}
		\end{split}
		\label{eqn:suploss}
	\end{equation}
	
	where i and j correspond to the feature coordinates, $\eta$ is a scalar hyper-parameter, and {\footnotesize $ \ce(\gtmask, \predmask) = - \frac{1}{W \times H} \sum_{i, j}\gtmask(i, j) \log(\predmask (i, j)) + (1 - \gtmask (i, j)) \log(1 - \predmask (i, j))$}. Note that this loss is computed both for positive pairs (source and target pairs generated by segment swapping) and negative pairs (sampled from two different pairs, without repeated objects) for which $\gtmask^s=\gtmask^t=0$ and by convention $\lcorr=0$.

	\paragraph{Implementation details}
	\label{sec:imple}
	We implement our approach using the Pytorch library. We use as backbone features the \emph{conv4} features of a ResNet-50~\cite{he2016deep} trained on ImageNet~\cite{deng2009imagenet} with MOCO-v2~\cite{chen2020improved}. 
	
	We freeze the backbone during the training, as learning backbone features leads to overfiting on the synthetic training set. For all the experiments, we optimise the loss defined in Equation~\ref{eqn:suploss} with $\eta=8$ and use the Adam optimiser~\cite{kingma2014adam} with momentum terms $\beta_1=0.5$ and $\beta_2=0.999$. At each iteration, we sample 5 positive and 15 negative pairs. For the transformer architecture, after training 200k iterations with learning rate 2e-4, we train with hard negative pairs and learning rate 1e-5  for 5k iterations. 
	Hard negatives are obtained by sampling a pool of $\negpool=500$ images from different synthetic pairs, computing predicted masks for all the pairs of images in the pool, and keeping those with mask prediction higher than a threshold $\tau$ = 0.04 in a hard negative pair pool for $\negiter=1000$ iterations of training.  
	For Sparse Nc-Net~\cite{rocco2020efficient} training 200k iterations with learning rate 2e-4 without hard negative mining leads to the best performance. The entire trainings of the transformer and Sparse Nc-Net~\cite{rocco2020efficient} take approximately 30 hours and 15 hours respectively on a single GPU Tesla-V100-16GB. An ablation study of the architectures and more training details are provided in the supplementary material~\cite{supp_mat}.

	\section{Application to Image Retrieval} 
	\label{sec:retrieval}
	In this section, we show how our model can be used for retrieval tasks. We first explain how we use it to compute an image similarity score in Section~\ref{sec:score}. {We then present experimental results in Section~\ref{sec:brueghel}, including art detail retrieval on the Brueghel dataset~\cite{shen2019discovering}, place recognition on Pitts30k~\cite{torii2013visual} and Tokyo 24/7~\cite{torii201524}, validation of our unsupervised segments and ablation study.} 
	More visual results and ablation studies are provided in the supplementary material~\cite{supp_mat}.
	
	\subsection{Score between a pair of images}
	\label{sec:score}
	We propose the following score $\score$ to measure the similarity between a pair of images based on predicted correspondences and masks. $\score$ is the sum of weighted local features similarities, where our predicted correspondences are used to associate features and the weight $\symmask^{s}$ is the product of the source and transported target mask {\footnotesize $\symmask^{s}(i,j) =\predmask^t(\predcorr^{s\rightarrow t}(i,j)) \predmask^s (i,j)$}: 
	
	\begin{equation}
	\scriptsize
		\begin{aligned} 
			\score(\img^s, \img^t) &= \sum_{i,j} \underbrace{\symmask^{s} (i,j)}_{\text{Mask}} \underbrace{cos (\feat^s(i,j), \feat^t (\predcorr^{s \rightarrow t}(i,j)))}_{\text{Feat. similarity}} 
		\end{aligned}
		\label{eqn:score}
	\end{equation}
	Ablations in supplementary material~\cite{supp_mat} show that the mask term is the key part of this score and that the feature similarity term provides an additional small boost.

	\subsection{Experiments}
	\label{sec:brueghel}
	Qualitative results on our different datasets can be seen in Figure~\ref{fig:vis}. The predicted masks, shown with transparency, are able to capture repeated regions even in challenging cases, such as large difference of scale, viewpoints, lightening conditions and depiction styles. More visual results are provided in the supplementary material~\cite{supp_mat}. 
	
	\begin{table}[!thbp]
    \begin{center}
		\resizebox{\columnwidth}{!}{
			\begin{tabular}{l|c|c}
				\hline \hline
				\multirow{2}{*}{Feat. + Methods}  & \multicolumn{2}{c}{mAP}  \\
				& Retrieval & Det.(IoU > 0.3)\\
				\hline
				Shen et al.~\cite{shen2019discovering} + cos~\cite{shen2019discovering}& 75.5 & 75.3 \\  
				Shen et al.~\cite{shen2019discovering} + discovery~\cite{shen2019discovering}& 76.6 & 76.4 \\
				MocoV2~\cite{chen2020improved} + cos~\cite{shen2019discovering}& 79.0 & 78.7 \\  
				MocoV2~\cite{chen2020improved} + discovery~\cite{shen2019discovering}& 80.8 & 79.6 \\
				
				\hline  
				
				\multicolumn{3}{l}{\textbf{Ours + Unsupervised segments}} \\
				
				\hspace{2mm}Transformer &  81.8 & 79.4\\
				
				\hspace{2mm} Sparse-Ncnet & 82.8 & 73.4 \\
				\multicolumn{3}{l}{\textbf{Ours + COCO segments~\cite{lin2014microsoft}}} \\
				
				\hspace{2mm} Transformer &  \bf 84.4 & \bf 81.8\\
				
				\hspace{2mm} Sparse-Ncnet &  83.3 & 73.7\\
				
				\hline \hline
		\end{tabular}}
		\vspace{-1mm}
		\caption{Art detail retrieval and detection on Brueghel~\cite{shen2019discovering}. For detection, we employ ArtMiner (Brueghel~\cite{shen2019discovering} + cos~\cite{shen2019discovering}) as a post-processing and reports results with IoU > 0.3~\cite{shen2019discovering}}
		\label{tab::brueghel}
	\end{center}
\end{table}
	\paragraph{Art detail retrieval} We evaluate our approach on the Brueghel dataset~\cite{brueghel,shen2019discovering} in Table~\ref{tab::brueghel}. Our score allows us to directly retrieve images from a selected query detail. To further compare with the detection performance in ArtMiner~\cite{shen2019discovering}, we crop a 320 $\times$ 320 patch around the predicted regions and use  ArtMiner~\cite{shen2019discovering} as a post-processing to obtain the bounding box prediction. 
	The correspondences are more accurate for the cross-image transformer which achieves much better results for detection. {We also observe that, in this benchmark, the performances with unsupervised segments are close to the ones using COCO~\cite{lin2014microsoft} instance annotations, which suggests that our approach does not depend on human annotations.}
	Note that the best performance of ArtMiner is obtained with a discovery score which is expensive to compute and involves multi-scale feature matching and RANSAC. Our approach is thus simpler, faster and more effective.

	\paragraph{Place recognition} 
	In Table~\ref{tab::placeReco} we compare our approach to state of the art for place recognition on the Pitts30k~\cite{torii2013visual} and Tokyo 24/7~\cite{torii201524} datasets.
	The descriptions of the datasets are in the supplementary material~\cite{supp_mat}. 
	
	We follow the standard evaluation protocol~\cite{sattler2012image,gronat2013learning,torii2013visual,arandjelovic2014dislocation,torii201524,ge2020self}. The query image is correctly localized if one of the top N retrieved
	database images is within d = 25 meters from the ground truth
	TUM coordinate of the query. The recall is then reported for N = 1, 5, 10. For
	Tokyo 24/7 we follow~\cite{torii201524,ge2020self} and perform spatial non-maximal suppression on ranked database images before evaluation. To enable fast evaluation, we follow PatchVlad~\cite{hausler2021patch} and evaluate our score on the top-100 images given by NetVLAD~\cite{arandjelovic2016netvlad}. Although our approach is not specifically designed for place recognition, it achieves performances comparable to Patch-NetVLAD~\cite{hausler2021patch} without RANSAC. 
	Note that the competing approaches either employ specific supervisions or more complicated process such as RANSAC, while our approach is trained only with our synthetic segment swapping data. Note that on this task where retrieving discriminative repeated regions is sufficient and correspondence accuracy is not critical, the Nc-Net architecture preforms better. {Similar to the Brueghel results, leveraging COCO~\cite{lin2014microsoft} annotated segments leads to superior performance. Training with unsupervised segments still leads to competitive results using the NC-Net. However, it gives clearly worst results using the transformer architecture on Tokyo 24/7. %
	}
	
	\begin{table}[!t]
	\begin{center}
		\resizebox{\columnwidth}{!}{
			\begin{tabular}{l|c|ccc|ccc}
				\hline \hline
				\multirow{2}{*}{Method}  & \multirow{2}{*}{Supervision} & \multicolumn{3}{c|}{Tokyo 24/7~\cite{torii201524}} &  \multicolumn{3}{c}{Pitts30k-test~\cite{torii2013visual}}  \\ & 
				& R@1 & R@5 & R@10 & R@1 & R@5 & R@10 \\ \hline
				AP-GEM~\cite{revaud2019learning,hausler2021patch} & Image location & 40.3 &  55.6 & 65.4 & 75.3 &  89.3  & 92.5 \\
				DenseVLAD~\cite{torii201524,hausler2021patch} & Image location & 59.4 & 67.3 & 72.1 & 77.7 & 88.3 &  91.6 \\
				NetVLAD~\cite{arandjelovic2016netvlad,hausler2021patch} & Image location & 73.3 & 82.9 & 86.0 & 86.0 & 93.2 & 95.1 \\
				CRN~\cite{kim2017learned,ge2020self} & Image location & 75.2 & 83.8 & 87.3 & - & - & -  \\
				SARE~\cite{liu2019stochastic,ge2020self} & Image location & 79.7 & 86.7 & 90.5 & - & - & - \\
				IBL~\cite{ge2020self} & Image location & \bf 85.4 & \bf 91.1 & \bf 93.3 & - & - & - \\
				\hline
				\multicolumn{8}{c}{\textbf{Re-ranking Top-100 from NetVLAD~\cite{arandjelovic2016netvlad,hausler2021patch}}} \\
				Patch-NetVLAD~\cite{hausler2021patch} & Image location & 81.9& 85.7& 87.9& 88.6& 94.5& 95.8\\
				
				Patch-NetVLAD~\cite{hausler2021patch} + RANSAC & Image location & 86.0& 88.6& \bf 90.5& \bf 88.7& 94.5& 95.9\\
				
				SuperGlue~\cite{sarlin2020superglue,hausler2021patch}$^{\star}$ & Pose+Depth& \bf 88.2 & \bf 90.2 & 90.2& \bf 88.7 & \bf 95.1 & \bf 96.4 \\
				
				\multicolumn{8}{l}{\textbf{Ours + Unsupervised segments}} \\
				\hspace{2mm} Transformer & \multirowcell{1}{Segment swapping}& 74.0 & 82.9 & 86.0 & 85.2 & 93.5 & 95.4  \\
				\hspace{2mm} Nc-Net & \multirowcell{1}{Segment swapping}& 84.1 & 87.0 & 88.9 & 86.4 & 94.3 & 95.6 \\
				\multicolumn{8}{l}{\textbf{Ours + COCO segments~\cite{lin2014microsoft}}} \\
				\hspace{2mm} Transformer & \multirowcell{1}{Segment swapping}& 80.0 & 86.0 & 87.9  & 84.7 & 93.5 & 95.6  \\
				\hspace{2mm} Nc-Net & \multirowcell{1}{Segment swapping}&  85.4 & 88.3 & 89.2  & 86.8 & 94.4 & 95.8  \\
				
				\hline \hline
		\end{tabular}}
		\vspace{-1mm}
		\footnotesize{$^{\star}$ uses learnt keypoint detector Superpoint~\cite{detone2018superpoint}}
		\caption{Image-based localization on Tokyo 24/7~\cite{torii201524} and Pitts30k~\cite{torii2013visual}. We follow Patch-NetVLAD~\cite{hausler2021patch} and re-rank the top-100 images ranked by NetVLAD~\cite{arandjelovic2016netvlad} features.}
		\label{tab::placeReco}
	\end{center}
	
\end{table} 
	\paragraph{Validation of our unsupervised segments}
{We evaluate our unsupervised segments for unsupervised saliency detection on three standard datasets ECSSD~\cite{shi2015hierarchical}, DUTS~\cite{wang2017learning} and DUT-OMRON~\cite{yang2013saliency}. The description of the datasets are provided in the supplementary material~\cite{supp_mat}. The results are shown in Table~\ref{tab:saliency}. Adding Bilateral Solver~\cite{barron2016fast} largely improve the performance over LOST~\cite{LOST} and achieves a significant boost on all the datasets compared to state-of-the-art approaches. We hope that this simple approach can serve as a new stronger baseline for unsupervised saliency detection. Qualitative results are provided in the supplementary material~\cite{supp_mat}.}

	\paragraph{Ablation study}
	\label{sec:abl}
	An ablation study of our approach using the cross-image transformer architecture and COCO~\cite{lin2014microsoft} annotated segments is shown in Table~\ref{tab::abl} on the Brueghel~\cite{brueghel,shen2019discovering} and Tokyo24/7~\cite{torii201524} datasets. We notice that: \emph{(i)} Poisson blending~\cite{perez2003poisson} and style transfer~\cite{huang2017arbitrary} are both critical;  \emph{(ii)} the three terms of the loss are necessary for good performance. Removing any of these elements results in very strong performance loss.
	More analysis on the importance of learning correspondences, the similarity score and the architectures are provided in the supplementary material~\cite{supp_mat}.
	
	\paragraph{Limitation} Our approach needs to compare pairs of images for retrieval, which makes it hard for large scale image retrieval applications.

    \section{Application to Object Discovery and Co-segmentation}
	In this section, we introduce an approach to use our predicted masks and correspondences for object discovery. We first explain how we designed a correspondences graph on which it is possible to perform spectral analysis in  Section~\ref{sec:cluster}. We then present experimental results on the dataset of~\cite{rubinstein2013unsupervised} for co-segmentation and Brueghel~\cite{brueghel,shen2019discovering} for discovery in Section~\ref{sec:graph_exp}.

		\begin{table}[!t]
	\begin{center}
		\resizebox{\linewidth}{!}{
			\begin{tabular}{cc|ccc|cc}
				\hline 
				\hline
				\multicolumn{2}{c|}{Dataset} & \multicolumn{3}{c|}{Losses} & \multicolumn{2}{c}{Cross-image Transformer} \\
				\multirow{2}{*}{Posson blending~\cite{perez2003poisson}} & \multirow{2}{*}{Style transfer~\cite{huang2017arbitrary}} & \multirow{2}{*}{$\lm$} &\multirow{2}{*}{$\ltm$} & \multirow{2}{*}{$\lcorr$} & Brueghel~\cite{brueghel,shen2019discovering} & Tokyo 24/7~\cite{torii201524} \\
				
				& & & & & mAP & R@1 \\
				\hline 
				$\cmark$ & $\cmark$ & $\cmark$ & $\cmark$ & $\cmark$ & \bf 84.4& \bf 80.0\\
				\hline 
				$\xmark$ & $\cmark$ & $\cmark$ & $\cmark$ & $\cmark$ & 75.1& 60.0 \\
				$\cmark$ & $\xmark$ & $\cmark$ & $\cmark$ & $\cmark$ & 75.6& 57.8 \\
				$\cmark$ & $\cmark$ & $\xmark$ & $\cmark$ & $\cmark$ & 80.9 & 67.8 \\
				$\cmark$ & $\cmark$ & $\cmark$ & $\xmark$ & $\cmark$ & 79.8 & 61.3 \\
				$\cmark$ & $\cmark$ & $\cmark$ & $\cmark$ & $\xmark$ & 8.5 & 13.3 \\
				
				\hline 
				\hline
				
		\end{tabular}}
		\vspace{-3mm}
		\caption{Ablation study. We report retrieval mAP on Brueghel~\cite{brueghel,shen2019discovering} and R@1 on Tokyo 24/7~\cite{torii201524} with our cross-image transformer using COCO segments~\cite{lin2014microsoft}.}
		\vspace{-5mm}
		\label{tab::abl}
	\end{center}
\end{table}

	\begin{table}[!t]
    \begin{center}
		\resizebox{\linewidth}{!}{
			\begin{tabular}{l|ccc|ccc|ccc}
				\hline 
				\hline
				\multirow{2}{*}{Method}  & \multicolumn{3}{c|}{ECSSD~\cite{shi2015hierarchical}} & \multicolumn{3}{c|}{DUTS~\cite{wang2017learning}} & \multicolumn{3}{c}{DUT-OMRON~\cite{yang2013saliency}}\\  
				& $\max F_\beta$ & IoU & Acc. & 
				$\max F_\beta$ & IoU & Acc. &
				$\max F_\beta$ & IoU & Acc. \\ \hline
			HS~\cite{yan2013hierarchical} & 0.673 & 0.508 & 0.847 & 0.504 & 0.369 & 0.826 & 0.561 & 0.433 & 0.843 \\
				wCtr~\cite{zhu2014saliency} & 0.684 & 0.517 & 0.862 & 0.522 & 0.392 & 0.835 & 0.541 & 0.416 & 0.838 \\
				WSC~\cite{li2015weighted} & 0.683 & 0.498 & 0.852 & 0.528 & 0.384 & 0.862 & 0.523 & 0.387 & \bf 0.865 \\
				DeepUSPS~\cite{nguyen2019deepusps} & 0.584 & 0.440 & 0.795 & 0.425 & 0.305 & 0.773 & 0.414 & 0.305 &  0.779 \\
				BigBiGAN~\cite{voynov2021object} & 0.782 & 0.672 & 0.899 & 0.608 & 0.498 & 0.878 & 0.549 & 0.453 & 0.856 \\
				E-BigBiGAN~\cite{voynov2021object} & 0.797 & 0.684 & 0.906 & 0.624 & 0.511 & 0.882 & 0.563 & 0.464 & 0.860 \\
				\hline 
				LOST~\cite{LOST} & 0.758 & 0.654 & 0.895 & 0.611 & 0.518 & 0.871 & 0.473 & 0.410 & 0.797 \\
				LOST~\cite{LOST} + Bilateral Solver~\cite{barron2016fast} (Ours) & \bf 0.837 & \bf 0.723 & \bf 0.916 & \bf 0.697 & \bf 0.572 & \bf 0.887 & \bf 0.578 & \bf 0.489 & 0.818   \\

				\hline 
				\hline   
		\end{tabular}}
		\caption{Unsupervised saliency detection on ECSSD~\cite{shi2015hierarchical}, DUTS~\cite{wang2017learning} and DUT-OMRON~\cite{yang2013saliency}. Our approach achieves clearly better results compared to competitive approaches.}
		\label{tab:saliency}
		
	\end{center}
			
\end{table} 
	\label{sec:graph}
	\subsection{Correspondences graph and clustering}
	\label{sec:cluster}
	In the spirit of~\cite{leordeanu2005spectral}, we see object discovery as a graph clustering problem, where the vertices $\vertice$ of the graph  $\graph$ = ($\vertice$, $\edge$) are correspondences between images and the weights of the edges encodes consistency between the correspondences.  Let us consider a set of $N$ images $(I_1, \ ...,\ I_N )$. For every pair of images %
	our network predicts correspondences that we add to the set of vertices $\vertice$ if the associated mask value is higher than a threshold. Each vertex 
	$v_i=(s_i,t_i,x^s_i,x^t_i,m_i)$ in the graph is thus associated to a predicted correspondence and defined by the indices $s_i$ and $t_i$ of the images it connects, the associated coordinates $x^s_i$ and $x^t_i$ and the predicted mask value $m_i$. We use the masks values and cycle consistency between the correspondences to define the weights of the edges between the different vertices. More precisely, we only connect correspondences which have exactly one image in common. For example, let's assume that we have two vertices $v_i$ and $v_j$ such that $s_i=s_j=s$ and $t_i\neq t_j$. We use our network to predict correspondence fields  $\predcorr^{t_i \rightarrow t_j}$ and $\predcorr^{t_j \rightarrow t_i}$ and we define the weight $\epsilon_{i,j}$ of the edge between  $v_i$ and $v_j$ as:
	
	\begin{equation}
	\scriptsize
		\begin{split}
			\epsilon_{i,j} & = \frac{1}{2} m_i m_j exp(-\frac{ \|x^s_i - x^s_j \|}{\sigma}) (exp(-\frac{ \| x^t_i - \predcorr^{t_j \rightarrow t_i}(x^t_{j})  \| }{\sigma}) \\ 
			& + exp(-\frac{\| x^t_j - \predcorr^{t_i \rightarrow t_j}(x^t_{i})  \| }{\sigma}))
		\end{split}
	\end{equation}
	
	where $\sigma$ is a scalar hyper-parameter. The edges are defined similarly in the cases $s_i=t_j$,  $t_i=s_j$ and $t_i=t_j$.
	More details about the way we define the graph and in particular strategies to limit the number of vertices are given in the supplementary material~\cite{supp_mat}.
	
		\begin{table}
		\begin{center}
			\resizebox{\columnwidth}{!}{
				\begin{tabular}{l|cc|cc|cc|cc}
					\hline 
					\hline
					\multirow{2}{*}{Method}  & \multicolumn{2}{c|}{Airplane} & \multicolumn{2}{c|}{Car} & \multicolumn{2}{c|}{Horse} & \multicolumn{2}{c}{Avg} \\  
					& $\mathcal{P}$ & $\mathcal{J}$ & $\mathcal{P}$ & $\mathcal{J}$ & $\mathcal{P}$ & $\mathcal{J}$ & $\mathcal{P}$ & $\mathcal{J}$ \\ \hline 
					DOCS~\cite{li2018deep}$^{*}$ & 0.946 & 0.64 & 0.940&  0.83 & 0.914 & 0.65 & 0.933 & 0.70 \\ \hline 
					Sun et al.~\cite{sun2016learning} & 0.886 & 0.36 & 0.870 & 0.73 & 0.876 & 0.55 & 0.877 & 0.55 \\
					Rubinstein et al. ~\cite{rubinstein2013unsupervised} & 0.880 & 0.56 & 0.854 & 0.64 & 0.828 & 0.52 & 0.827 & 0.43 \\
					Chen et al. ~\cite{chen2014enriching} & 0.902 & 0.40 & 0.876 & 0.65 & 0.893 & 0.58 & 0.890 & 0.54 \\
					Quan et al. ~\cite{quan2016object} & 0.910 & 0.56 & 0.885 & 0.67 & 0.893 & 0.58 & 0.896 & 0.60 \\
					Chang et al.~\cite{chang2015optimizing} & 0.726 & 0.27 &  0.759 & 0.36 & 0.797 & 0.36 & 0.761 & 0.33 \\
					Lee et al.~\cite{lee2015multiple} & 0.528& 0.36 &0.647 &0.42 &0.701& 0.39& 0.625& 0.39 \\
					Jerripothula et al.~\cite{jerripothula2016image} & 0.905 & 0.61 & 0.880 & 0.71 & 0.883 & 0.61 & 0.889 & 0.64 \\
					Hsu et al.~\cite{hsu2018co} & 0.936 &  0.66 & 0.914 & 0.79 & 0.876 & 0.59 & 0.909 & 0.68 \\
					Chen et al.~\cite{chen2020show} &  \bf 0.941 & 0.65 & \bf 0.940 & \bf 0.82 & \bf 0.922 & \bf 0.63 & \bf 0.935 & \bf 0.70 \\
					\hline 
					\multicolumn{9}{l}{\textbf{Ours + Unsupervised segments}} \\
					\hspace{2mm} transformer & 0.925 & 0.65 & 0.914 & 0.79 & 0.909 & 0.60 & 0.916 & 0.68 \\
					\hspace{2mm} Nc-Net & 0.746 & 0.25 & 0.874 & 0.68 & 0.836 & 0.38 & 0.819 & 0.44  \\
					\multicolumn{9}{l}{\textbf{Ours + COCO segments~\cite{lin2014microsoft}}} \\
					\hspace{2mm} transformer & \bf 0.941& \bf 0.67& 0.928 &\bf 0.82 & 0.916 & 0.60 & 0.928 & \bf 0.70 \\
					\hspace{2mm} Nc-Net  & 0.655 &0.23 & 0.857& 0.61&0.873 & 0.43& 0.795& 0.42\\
					\hline 
					\hline   
					
			\end{tabular}}
			\footnotesize{$^{*}$ learned with strong supervision (i.e., manually annotated object masks)}
			\caption{Co-segmentation on the dataset of~\cite{rubinstein2013unsupervised}. We report pixel level precision $\mathcal{P}$ and Jaccard index $\mathcal{J}$}
			\label{tab:co-seg}
			
		\end{center}
		
	\end{table}
	Given the correspondence graph, we use the spectral decomposition of its adjacency matrix~\cite{ng2001spectral,leordeanu2005spectral} either to obtain clusters of correspondences for object discovery, or a foreground potential for co-segmentation. For discovery we first compute $\nbeig$ principal eigenvectors then performing K-means with $\nbcluster$ clusters. For co-segmentation, we directly use the first eigenvector to define a foreground potential. Note that because we only consider in the graph correspondences with mask values higher than a threshold, the full graph is extremely sparse that the eigen-decomposition can be efficiently computed. 
	
	\begin{figure*}[!thbp]
		\centering
		\subfloat[Co-segmentation results in the dataset of ~\cite{rubinstein2013unsupervised} for the  Horse, Airplane and Car categories.]{\includegraphics[width=0.9\textwidth]{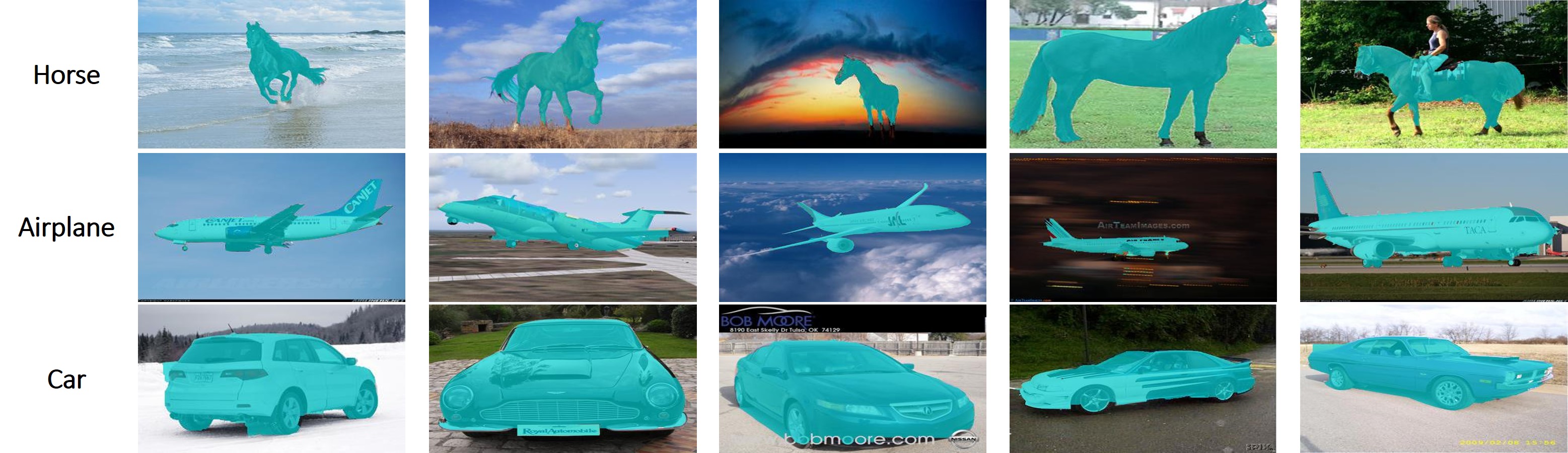}\label{fig:int}} \hfill
		\subfloat[Examples of discovered clustered on Brueghel~\cite{brueghel,shen2019discovering}.]{\includegraphics[width=0.9\textwidth]{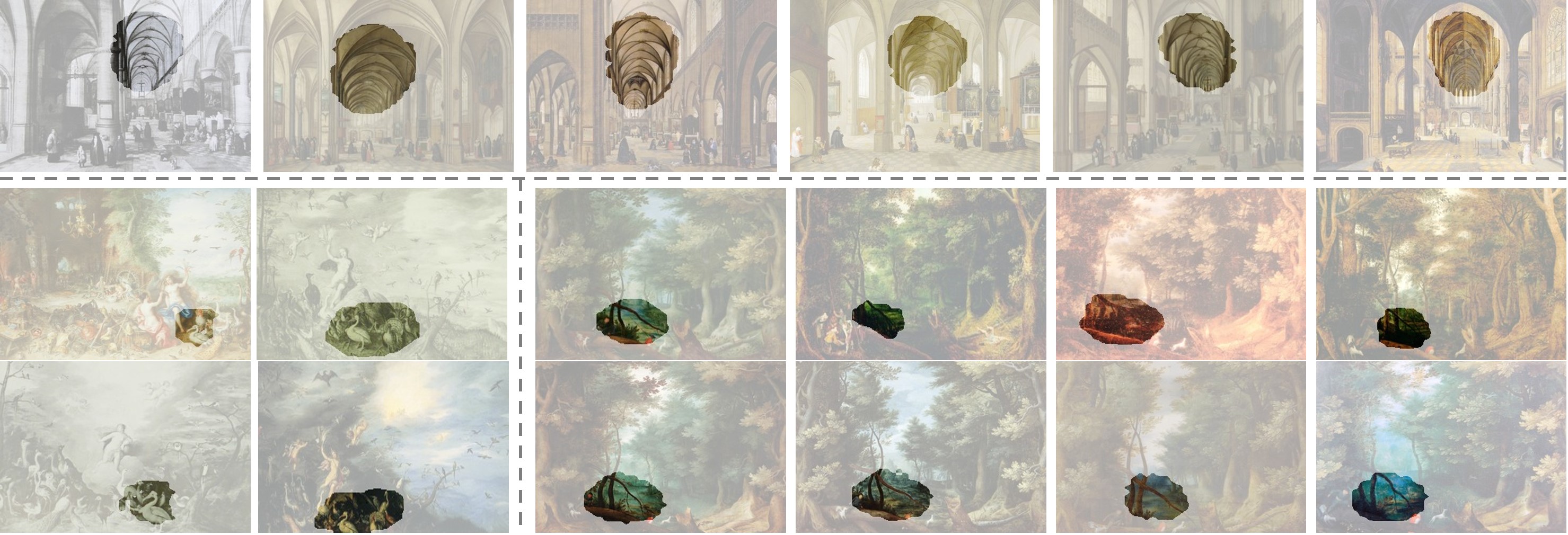}\label{fig:vis_brueghel}} 
		\caption{Visual results for object discovery. We show co-segmentation results on the dataset of~\cite{rubinstein2013unsupervised} in Figure~\ref{fig:int} and some discovered clusters in Brueghel~\cite{shen2019discovering,brueghel} in Figure~\ref{fig:vis_brueghel}.}
		\label{fig:vis_dis}
		
	\end{figure*}
	
	\subsection{Experiments}
	\label{sec:graph_exp}

	\paragraph{Object co-segmentation on the dataset of~\cite{rubinstein2013unsupervised}} 
	We build the correspondences graph using for each image only the correspondences in the five most similar images according to the retrieval score of Equation~\ref{eqn:score}. 
	We then use the principal eigen-vector of the correspondence graph to define a seed for GrabCut~\cite{rother2004grabcut}. More precisely, for every image we associate to each position the sum of the eigen-vector values for the correspondences at this position. Note that GrabCut~\cite{rother2004grabcut} is crucial to achieve good performance on this dataset, and is widely used by competing approaches such as~\cite{rubinstein2013unsupervised,jerripothula2016image,quan2016object,hsu2018co,chen2020show}. More details about the GrabCut~\cite{rother2004grabcut} can be found in the supplementary material~\cite{supp_mat}. We follow the standard evaluation protocol~\cite{rubinstein2013unsupervised,chen2020show} and report pixel-level precision $\mathcal{P}$ and the Jaccard index $\mathcal{J}$ on three subsets: Airplane, Car, Horse. The precision $\mathcal{P}$ measures pixel accuracy. The Jaccard index $\mathcal{J}$ is the IoU between the segmented object and ground truth object. Quantitative results are presented in Table~\ref{tab:co-seg} and qualitative results in  Figure~\ref{fig:int}. Our cross-image transformer obtains performance comparable to the state of the art unsupervised approaches. {Again, the performances using annotated COCO~\cite{lin2014microsoft} segments and unsupervised segments are close, which demonstrates that the success of our approach does not come from implicitly leveraging annotated object segmentations}. Sparse Nc-Net performances are clearly worse for this task. This can be understood by looking at qualitative results: the segmentation masks predicted by Nc-Net tend to be more localized in discriminative regions.
	
	\paragraph{Discovery on Brueghel dataset~\cite{brueghel,shen2019discovering}} To compute correspondences and build our correspondence graph, we first resize all images to 640 $\times$ 640, as many repeated details in Brueghel~\cite{brueghel,shen2019discovering} are small. We also remove duplicate images and images with similar borders, easily detected with our algorithm, to focus on more interesting repeated details. 
	Again, we only include in the graph the correspondences from the five most similar images according to the retrieval score to limit the size of the graph and we perform K-means for $\nbcluster=500$ clusters with $\nbeig = 100$ principal eigen vectors. 
	Figure~\ref{fig:vis_brueghel} presents some interesting clusters that are not covered by ArtMiner~\cite{shen2019discovering}\footnote{\scriptsize{ \url{http://imagine.enpc.fr/~shenx/ArtMiner/visualRes/brueghel/brueghel.html}}}. More results and details are  in the supplementary material~\cite{supp_mat}.
	
	\paragraph{Limitation} The current version of our approach is not scalable to large datasets, as the dimensions of the graph is quadratic on number of images. For example, to perform the discovery in Brueghel~\cite{brueghel,shen2019discovering}, the graph consists of $\sim$900K nodes and it took 10 hours to compute predictions of all the pairs and 2 hours to perform the eigen-decomposition and clustering.
	
	\section{Conclusion} 
	In this work, we presented a ``segment swapping'' approach to generate pairs of images with repeated patterns from which we show it is possible to train networks to predict co-segmentation. We evaluated two architectures, a cross-image transformer we introduced and a modified Sparse Nc-Net~\cite{rocco2020efficient}. {We also compared using annotated segments in COCO~\cite{lin2014microsoft} and segments extracted in a completely unsupervised way, which shows that our approach is not reliant on COCO~\cite{lin2014microsoft} object annotations.} We demonstrated the interest and generality of the trained co-segmentation networks by showing competitive or better performance compared to specialized baseline on a wide range of datasets and tasks, including art detail retrieval, place recognition and object discovery. 
	
	\paragraph{Acknowledgement} This work was supported in part by ANR project EnHerit ANR-17-CE23-0008, project Rapid Tabasco, and IDRIS under the allocation AD011011160R1 made by GENCI.
	{\small
		\bibliographystyle{ieee_fullname}
		\bibliography{egbib}
	}
	
\end{document}